\documentclass[sigconf]{aamas} 

\usepackage{balance} 
\usepackage{amsmath}
\usepackage{subcaption}
\usepackage{algorithm}
\usepackage{algpseudocode}
\usepackage{multirow}
\usepackage{booktabs}

\doi{CMRP6518}

\makeatletter
\gdef\@copyrightpermission{
  \begin{minipage}{0.2\columnwidth}
   \href{https://creativecommons.org/licenses/by/4.0/}{\includegraphics[width=0.90\textwidth]{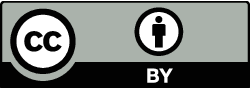}}
  \end{minipage}\hfill
  \begin{minipage}{0.8\columnwidth}
   \href{https://creativecommons.org/licenses/by/4.0/}{This work is licensed under a Creative Commons Attribution International 4.0 License.}
  \end{minipage}
  \vspace{5pt}
}
\makeatother

\setcopyright{ifaamas}
\acmConference[AAMAS '26]{Proc.\@ of the 25th International Conference
on Autonomous Agents and Multiagent Systems (AAMAS 2026)}{May 25 -- 29, 2026}
{Paphos, Cyprus}{C.~Amato, L.~Dennis, V.~Mascardi, J.~Thangarajah (eds.)}
\copyrightyear{2026}
\acmYear{2026}
\acmDOI{}
\acmPrice{}
\acmISBN{}

\title[AAMAS-2026 Formatting Instructions]{Sim2Sea: Sim-to-Real Policy Transfer for Maritime Vessel Navigation in Congested Waters}

\author{Xinyu Cui}
\affiliation{
  \institution{Institute of Automation, CAS}
  \institution{School of Artificial Intelligence, UCAS}
  \city{Beijing}
  \country{China}}
\email{cuixinyu2021@ia.ac.cn}

\author{Xuanfa Jin}
\affiliation{
  \institution{Institute of Automation, CAS}
  \institution{School of Artificial Intelligence, UCAS}
  \city{Beijing}
  \country{China}}
\email{jinxuanfa2022@ia.ac.cn}

\author{Xue Yan}
\affiliation{
  \institution{Institute of Automation, CAS}
  \institution{School of Artificial Intelligence, UCAS}
  \city{Beijing}
  \country{China}}
\email{yanxue2021@ia.ac.cn}

\author{Yongcheng Zeng}
\affiliation{
  \institution{Institute of Automation, CAS}
  \city{Beijing}
  \country{China}}
\email{zengyongcheng2022@ia.ac.cn}

\author{Luoyang Sun}
\affiliation{
  \institution{Institute of Automation, CAS}
  \city{Beijing}
  \country{China}}
\email{sunluoyang2022@ia.ac.cn}

\author{Siying Wei}
\affiliation{
  \institution{Institute of Automation, CAS}
  \city{Beijing}
  \country{China}}
\email{weisiying2022@ia.ac.cn}

\author{Ruizhi Zhang}
\affiliation{
  \institution{Institute of Automation, CAS}
  \city{Beijing}
  \country{China}}
\email{ruizhi.zhang@ia.ac.cn}

\author{Jian Zhao}
\affiliation{
  \institution{Zhongguancun Academy}
  \city{Beijing}
  \country{China}}
\email{jianzhao@zgci.ac.cn}

\author{Haifeng Zhang}
\authornote{Corresponding to Haifeng Zhang <haifeng.zhang@ia.ac.cn>.}
\affiliation{
  \institution{Institute of Automation, CAS}
  \city{Beijing}
  \country{China}}
\email{haifeng.zhang@ia.ac.cn}

\author{Jun Wang}
\affiliation{
  \institution{University College London}
  \city{London}
  \country{United Kingdom}}
\email{jun.wang@cs.ucl.ac.uk}

\begin{abstract}

Autonomous navigation in congested maritime environments is a critical capability for a wide range of real-world applications. However, it remains an unresolved challenge due to complex vessel interactions and significant environmental uncertainties. Existing methods often fail in practical deployment due to a substantial sim-to-real gap, which stems from imprecise simulation, inadequate situational awareness, and unsafe exploration strategies. To address these, we propose \textbf{Sim2Sea}, a comprehensive framework designed to bridge simulation and real-world execution. 
Sim2Sea advances in three key aspects. First, we develop a GPU-accelerated parallel simulator for scalable and accurate maritime scenario simulation. Second, we design a dual-stream spatiotemporal policy that handles complex dynamics and multi-modal perception, augmented with a velocity-obstacle-guided action masking mechanism to ensure safe and efficient exploration. 
Finally, a targeted domain randomization scheme helps bridge the sim-to-real gap.
Simulation results demonstrate that our method achieves faster convergence and safer trajectories than established baselines. In addition, our policy trained purely in simulation successfully transfers zero-shot to a 17-ton unmanned vessel operating in real-world congested waters. These results validate the effectiveness of Sim2Sea in achieving reliable sim-to-real transfer for practical autonomous maritime navigation.

\end{abstract}


\newcommand{\BibTeX}{\rm B\kern-.05em{\sc i\kern-.025em b}\kern-.08em\TeX}

\begin{document}


\pagestyle{fancy}
\fancyhead{}


\maketitle 


\section{Introduction}
As global maritime trade intensifies and nautical activities in near-shore congested waters expand, autonomous navigation is becoming critical for enhancing safety and operational efficiency. These environments, such as ports and coastal fairways, are notoriously difficult to navigate due to dense, heterogeneous traffic, complex geographical constraints, and unpredictable environmental forces. This complexity often leads to scenarios where traditional navigation methods fall short. While rule-based systems grounded in the Convention on the International Regulations for Preventing Collisions at Sea (COLREGs) or collision avoidance algorithms like Velocity Obstacles (VO) provide a crucial safety baseline~\cite{naeem2012colregs, maza2022colregs}, there are still inherent limitations. The ambiguity of rule-based methods in multi-vessel encounters and their inadequacy in handling mixed obstacle types (e.g., ships and static infrastructure) can lead to indecisive or overly conservative actions~\cite{kuwata2013safe, zhou2020study}. This motivates the development of adaptive, learning-based methods capable of flexible decision-making amid variable traffic dynamics.

Reinforcement learning (RL)~\cite{kaelbling1996reinforcement, li2017deep} has emerged as a promising approach for autonomous nautical operations~\cite{zhao2019colregs, fan2022novel}. However, the path to deploying RL in the real world is fraught with significant obstacles. The foremost challenge is the lack of suitable training environments. To our knowledge, there is no open-source, high-performance simulator that can accurately model the complex dynamics and diverse interaction scenarios required for training a capable maritime agent.

Furthermore, even with a hypothetical perfect simulator, several challenges remain. First, agents should navigate using a multi-modal and asynchronous stream of data from sources like Automatic Identification System (AIS) signals, radar signals, and nautical charts, making the creation of a coherent environmental understanding a significant perceptual hurdle. Second, the substantial inertia and underactuated nature of vessels introduce complex temporal dynamics. Purely reactive policies might fail as they cannot account for momentum or external forces like currents. Finally, policies trained in even the most detailed simulations often fail when transferred to a physical vessel due to the sim-to-real gap caused by subtle but critical differences in dynamics, sensor noise, and actuation delays, which is a major challenge to practical deployment.

To address these challenges, we present \textbf{Sim2Sea}, an integrated framework designed to enable the development and real-world deployment of maritime autonomous navigation agents. 
Our approach is built upon three core pillars.
First, we develop a high-performance parallel simulator with realistic vessel dynamics and continuous-time safety checks.
Second, we designed a dual-stream spatiotemporal policy that leverages a Transformer to understand temporal dynamics and a Bird's-Eye-View (BEV) image to process spatial context.
This policy is also augmented by a VO-guided action masking mechanism to ensure safe and efficient exploration. Third, we explicitly bridge the sim-to-real gap through targeted domain randomization, ensuring the learned policy is robust to real-world variability.
Our main contributions are as follows:
\begin{itemize}
    \item We introduce a high-speed parallel maritime simulator designed specifically for large-scale RL training, providing a powerful tool for the research community.
    \item We propose an agent architecture combining a spatiotemporal policy with active action masking, which is shown to be highly effective for stable and safe learning in complex maritime scenarios.
    \item 
    Trained with a targeted domain randomization scheme, our learning-based policy achieves successful zero-shot deployment on a 17-ton unmanned vessel in open-sea, congested waters. To the best of our knowledge, this is the first successful trial of its kind on a vessel of this scale.
\end{itemize}

\section{Related Works}
\subsection{Maritime Vessel Simulator}
The domain of maritime simulation has predominantly focused on Unmanned Underwater Vehicles (UUVs), with notable open-source and high-performance examples like UUVSim~\cite{zhang2024uuvsim} and DAVE~\cite{zhang2022dave}. However, these simulators focus on underwater scenarios. 
While proprietary simulators often use established hydrodynamic models like the Maneuvering Modeling Group (MMG)~\cite{yasukawa2015introduction, mmgdynamics} or Abkowitz~\cite{abkowitz1980measurement} for maritime vessels, they are typically not open-source and are poorly optimized for large-scale RL training.

Recent simulators like AquaNav~\cite{mane2025aquanav} prioritize visual fidelity at the cost of parallel processing and accurate vessel dynamics, whereas platforms like MARUS~\cite{lonvcar2022marus} and SMaRCSim~\cite{kartavsev2025smarcsim} lack a specific focus on ship characteristics. To address these deficiencies, we propose a parallel simulator designed for maritime vessels. It supports multiple modeling approaches for maritime vessels, such as the MMG and Nomoto models~\cite{golikov2018simple}, to effectively facilitate RL training. Furthermore, it features comprehensive modeling of sensors, complex marine environments, and dynamic sea states.

\subsection{Autonomous Maritime Navigation}
Autonomous maritime navigation has traditionally been dominated by rule-based implementations of COLREGs for Maritime Autonomous Surface Ships, which provide interpretable behavior but struggle in dynamic, congested, or unstructured environments~\cite{statheros2008autonomous, zhang2021collision}. To address these limitations, recent learning-based methods incorporate domain priors while optimizing decision-making under uncertainty. Adaptive systems that couple classical guidance or control with geometric collision-avoidance, such as fuzzy PID with enhanced velocity obstacles, improve multi-ship encounters by exploiting VO structure~\cite{zhao2025adaptive, fiorini1998motion}. RL also has been used to learn collision avoidance policies that respect COLREGs while scaling to complex interactions~\cite{pan2025deep}. To accelerate training and stabilize policy learning, several works integrate prior knowledge into RL and adjust policy entropy to balance safety and exploration under COLREGs constraints~\cite{zhang2025advancing, wang2023collision, chen2024novel}.

Despite this progress, most efforts that combine ship control with navigation focus on reward shaping grounded in VO or COLREGs, and systematic exploration near complex coastlines, ports, and restricted waterways remains limited. Existing studies often emphasize path planning rather than closed-loop ship control in such settings~\cite{kim2024advancing}. Motivated by these gaps, we aim to explore complex terrains and scenes with a control-oriented approach, and to more deeply leverage VO and related structure to assist RL training and online decision-making. 

\subsection{Sim-to-Real for Navigation Tasks}
Sim2real for navigation has progressed rapidly in autonomous driving, aerial vehicles, and mobile robots through a combination of domain randomization, domain adaptation, system identification, and policy distillation~\cite{hofer2021sim2real}. Recent driving and embodied navigation works couple world-model or BEV-based intermediate representations~\cite{liu2023world} with student–teacher transfer to bridge noisy sensing and actuation delays, while emphasizing simulation speed and robustness rather than only high visual fidelity to improve transfer reliability~\cite{truong2023rethinking}. Benchmarks and hybrid frameworks~\cite{haoran2023neuronsgym,azzam2022learning,bono2024learning} demonstrate that policies trained entirely in simulation can zero-shot or few-shot transfer to real robots when actuation dynamics, sensing noise, and goal specification are modeled consistently, with emerging results across navigation scenarios~\cite{chen2024sim2real}.

For surface vessels, sim2real studies are comparatively sparse, and most reported successes occur in constrained or lab-scale environments rather than open sea. One study applies deep RL with sim2real ideas to static obstacle avoidance by increasing simulation stochasticity but does not perform real-ship deployment~\cite{han2025robust}. Another line of work demonstrates sim-to-real transfer for small-scale platforms, such as achieving higher speed and energy efficiency for a small surface vessel~\cite{batista2024deep}, mapless navigation for a spherical micro-robot in a tank~\cite{wang2022sim}, or control for a wind-powered boat in a basin~\cite{bink2024autonomous}.

Current domain challenges include the lack of high-throughput simulators that can accurately model hydrodynamics, and the difficulty of bridging the sim-to-real gap for underactuated vessels. Our work addresses this by first building a high-performance parallel simulator to capture these complex dynamics. Subsequently, we utilize a spatiotemporal network and targeted domain randomization to achieve successful sim-to-real transfer for open-sea navigation on a full-sized vessel.

\section{Preliminaries}

\textbf{Problem Formulation.} Our work focuses on the problem of autonomous vessel navigation in congested near-shore waters. The primary objective is to develop an agent capable of guiding the vessel to a designated destination while ensuring collision-free operation with respect to both static geographical features and dynamic obstacles. Therefore, to facilitate agent training and policy optimization, we formulate this navigation task as a Partially Observable Markov Decision Process (POMDP)~\cite{krishnamurthy2016partially}, defined by the tuple $\langle \mathcal{S}, \mathcal{A}, \Omega, P, O, R \rangle$, where $\mathcal{S}$, $\mathcal{A}$, and $\Omega$ represent the state space, action space, and observation space, respectively. $P: \mathcal{S} \times \mathcal{A} \to \Delta(\mathcal{S})$ is the transition function, indicating the transition probability of taking actions at each state. $O: \mathcal{S} \to \Delta(\Omega)$ is the observation function, which describes what information the agent can obtain at each state. And $R: \mathcal{S} \times \mathcal{A} \to \mathbb{R}$ is the reward function, providing an immediate reward $r_t$ at each step $t$ given a state-action pair. Specifically, in our task settings, the state $s \in \mathcal{S}$ includes the vessel's details along with the positions and velocities of all surrounding obstacles. The observations $o \in \mathcal{O}$ are synthesized from simulated nautical charts, Global Navigation Satellite System (GNSS), AIS, and radar, providing the agent with partial information about the environment. And the action space $\mathcal{A}$ is discrete, consisting of eighteen commands for heading changes distributed uniformly over the interval $[-\pi, \pi]$.

At each step, the agent selects a desired heading, and the simulator propels the vessel toward this heading. An episode concludes when the agent successfully reaches the goal, a collision with an obstacle is registered by continuous-time collision detection, or the maximum number of steps is exceeded. The reward function is designed to foster time-efficient and safe navigation by rewarding progress toward the goal while imposing significant penalties for collisions and boundary violations.

\textbf{Proximal Policy Optimization (PPO).} To solve the described POMDP, we utilize PPO, an actor-critic reinforcement learning algorithm that is effective for complex control tasks. The primary objective for the policy network  is given by:
\begin{equation}
L^{CLIP}(\theta) = \hat{\mathbb{E}}_t \left[ \min\left(\rho_t(\theta)\hat{A}_t, clip(\rho_t(\theta), 1-\epsilon, 1+\epsilon)\hat{A}_t\right) \right].
\end{equation}
Here, $ \rho_t(\theta) $ represents the probability ratio between the new and old policies, $ \hat{A}_t $ is an estimator of the advantage function at time step $ t $, and $ \epsilon $ is a hyperparameter that dictates the clipping range.

Our implementation of PPO is tailored for the maritime vessel navigation task, incorporating features such as multimodal observations and state-dependent action masking. These specific adaptations and other implementation details will be discussed in the Methodology section.

\section{Methodology}

\begin{figure*}[ht]
    \centering
    \includegraphics[width=0.9\linewidth]{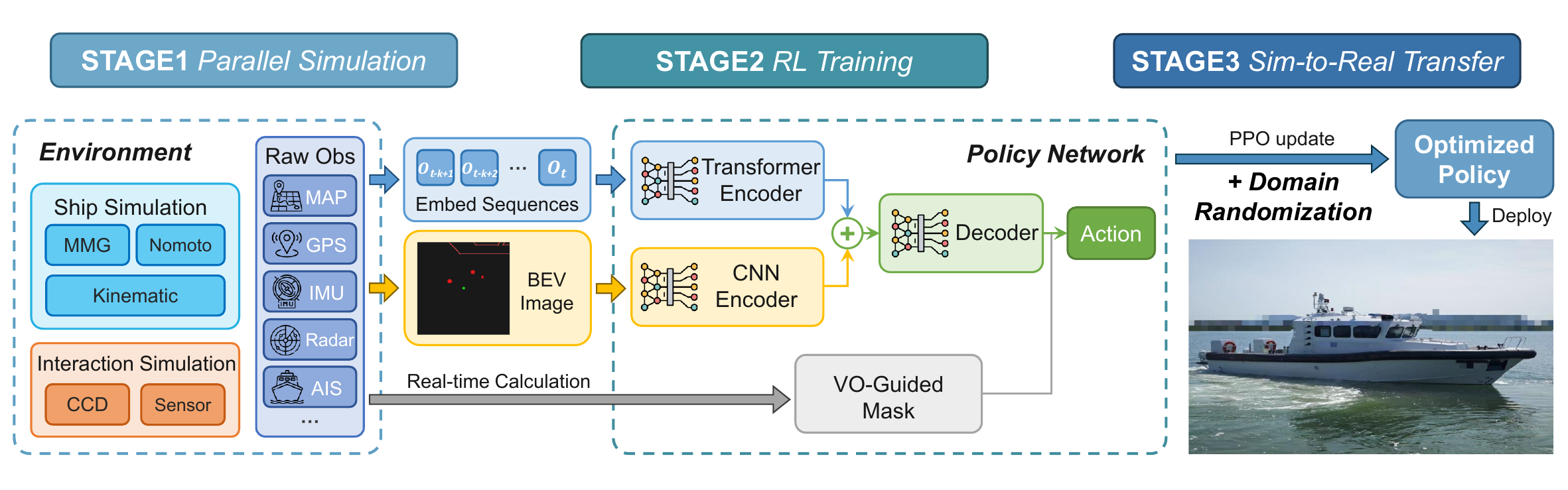} 
    \caption{Sim2Sea overview. Left: Parallel maritime simulator with multiple vessel models and interaction simulations. Middle: Spatiotemporal policy with BEV fusion and VO guided active action masking. Right: Zero-shot onboard deployment enabled by domain randomization.} 
    \label{fig:archi}
    \vskip -0.15in
\end{figure*}

In this section, we present Sim2Sea, our integrated framework for sim-to-real maritime vessel navigation. As illustrated in Figure~\ref{fig:archi}, our framework comprises three core components: (1) a high-performance parallel simulator for large-scale training; (2) a spatiotemporal decision network that fuses historical data with spatial context for robust control; and (3) a domain randomization strategy to ensure seamless transfer to real-world vessels. We begin by introducing the simulator, which serves as the foundation for policy learning, followed by a detailed description of the decision-making agent's architecture, and finally the sim-to-real transfer techniques.

\subsection{Maritime Vessel Navigation Simulator}

Inspired by the success of massively parallel simulators in fields like autonomous driving and robotics, we develop a high-throughput simulation environment tailored for maritime vessel navigation. The Sim2Sea's simulator is built on three design principles: high-fidelity modeling, high-performance parallelization, and efficient interaction computation.

\textbf{Fidelity and Flexibility in Ship Modeling.}
Sim2Sea distinguishes itself from existing simulators that often rely on simplified kinematics or general-purpose physics engines by providing specialized and physically-grounded vessel dynamics. It supports three distinct motion models to balance fidelity and computational cost. The primary model is a 3 degrees-of-freedom Maneuvering Modeling Group (MMG) dynamics implementation; the MMG equations are given by:
\begin{equation}
\begin{aligned}
m(\dot{u} - vr) &= X_H(u,v,r) + X_P(\cdot) + X_R(\cdot), \\
m(\dot{v} + ur) &= Y_H(u,v,r) + Y_P(\cdot) + Y_R(\cdot), \\
I_z \dot{r} &= N_H(u,v,r) + N_P(\cdot) + N_R(\cdot),
\end{aligned}
\end{equation}
where $m$ is the vessel mass and $I_z$ is its moment of inertia about the z-axis. The variables $u$, $v$, and $r$ are the surge, sway, and yaw velocities in the body-fixed frame, with $\dot{u}$, $\dot{v}$, and $\dot{r}$ as their respective accelerations. The terms $X$, $Y$, and $N$ represent forces and a moment, where the subscripts $H$, $P$, and $R$ denote contributions from hydrodynamics, the propeller, and the rudder. 
For applications demanding less detail, we also provide a Nomoto yaw-response model and a lightweight nonlinear kinematic model. To ensure numerical stability, the simulator supports both semi-implicit Euler and fourth-order Runge-Kutta (RK4) integration, with internal sub-stepping to handle accurate transients.

To align with real-world ship control systems, Sim2Sea not only supports direct maneuvering, but also abstracts low-level commands by incorporating PID controllers. This provides a unified, high-level control interface where the agent can directly command a target speed and heading, simplifying the policy learning task.

\textbf{High-Performance Parallelization.} 
While specialized simulators focus on detailed maneuvering analysis, they typically lack support for parallel execution and RL-centric interfaces. To address this problem, Sim2Sea's simulator is built using the Taichi language~\cite{hu2019taichi}, enabling native execution on both CPU and GPU backends. We employ an agent-centric parallelization strategy: for a configuration of $N$ environments with $M$ agents each, we execute a single 'struct-for' kernel over the entire set of $N \times M$ agents. This approach avoids serial computation within each environment, ensuring efficient scaling and high throughput, which are crucial for large-scale RL experiments.

\textbf{Efficient Interaction Computation.} 
Simulating interactions between vessels and with complex geography is computationally intensive. Our environment supports the import of both circular obstacles, representing other vessels from AIS or radar data, and polyline obstacles, representing geographical features like coastlines and breakwaters. Recognizing that interactions are spatially sparse even in congested areas, we use a hash-grid method for broad-phase collision detection to quickly identify potential collisions, significantly reducing unnecessary pairwise checks. 

For safety-critical evaluation, we employ continuous-time collision detection (CCD), which checks for intersections along the vessel's entire swept path between discrete steps. This mitigates the risk of missing collisions that can occur with simple discrete-time checks.


\subsection{Maritime Decision Network}

Our decision network is designed to handle the complexities of maritime environments, such as a variable number of dynamic obstacles and complex coastal geometries. The architecture consists of three key modules: a temporal encoder for understanding vessel dynamics, a spatial encoder for situational awareness, and an innovative action masking mechanism for ensuring safe exploration.

\textbf{Embedding of the environment and temporal sensing.} 
To address the challenge of variable-sized observations from dynamic obstacles~\cite{karamouzas2017implicit}, we use an implicit encoding scheme inspired by potential fields. Gradients derived from a logarithmic barrier function are aggregated to form a compact, fixed-size feature vector representing environmental risks.  
For agent at position $\mathbf{p}$ and an obstacle at distance $d(\mathbf{p})$ with safety radius $d_0$, the potential energy is defined as $U(d) = \alpha \log(d/d_0)(d_0 - d)^2$, where $\alpha$ is a tunable strength coefficient. The corresponding gradient is:
\begin{equation}
    \nabla U(\mathbf{d}) = \alpha \left(2 \log(d/d_0) \cdot (d - d_0) + (d - d_0) (1 - d_0/d) \right).
\end{equation}

We aggregate and normalize gradients from nearby obstacles, computing separate gradients for polyline obstacles, circular obstacles, and a goal-oriented composite term. This encoding compresses high-dimensional, variable-length obstacle data into a compact feature set that captures environmental risk without explicit enumeration. We also integrate the agent state, goal conditions, and other features with the gradients to form the observation $o_t$.

Since single-step observations are insufficient to capture vessel dynamics under currents and other disturbances, we utilize a transformer encoder to process a sequence of historical observations over the last $k$ time steps, represented as $\mathbf{o}_{t-k+1:t} \in \mathbb{R}^{k \times d}$, where $d$ is the state dimension. This temporal network captures long-range dependencies and recovers underlying environmental dynamics, making the decision more reliable. The temporal feature can be described as: $\mathbf{h}_{\text{temporal}} = \text{Transformer Encoder}(\mathbf{o}_{t-k+1:t})$.


\textbf{Enhancing the spatial awareness.} 
Simultaneously, to provide the agent with a comprehensive situational spatial picture, we adopt a Bird's Eye View (BEV) representation to encode spatial context around the agent vessel. Unlike terrestrial autonomous vehicles that construct BEV maps through computationally intensive fusion of camera and LiDAR data, maritime platforms benefit from radar and Automatic Identification System (AIS), which directly provide positions of surrounding vessels, and from nautical charts that supply dense geometric information on coastlines and navigational hazards. Leveraging the coordinate information from radar, AIS, and nautical charts, the BEV is constructed through an agent-relative rendering pipeline that incorporates dynamic scaling and depth ordering. This approach enables efficient and robust BEV generation, circumventing the computational complexity and reliability issues of multi-sensor fusion, especially under challenging conditions such as fog or severe vessel motion.


Concurrently, the BEV image is processed into a spatial representation $\mathbf{h}_{\text{spatial}}$ using a lightweight convolutional neural network (CNN) encoder.
Finally, temporal and spatial representations are fused and passed through an MLP decoder to generate the actor network output. The decoder produces action logits $z$ over the discrete action space as
\begin{equation}
z = \text{softmax}\big(\text{Decoder}([\mathbf{h}_{\text{temporal}};\mathbf{h}_{\text{spatial}}])\big).
\end{equation}
This dual-stream architecture enables robust decision-making in congested and dynamic maritime environments.

\textbf{Velocity Obstacles Guided Active Action Mask Generation.} 
While our temporal-spatial network enhances the agent's perception, exploration in geometrically complex environments can still lead to unsafe actions. 
Unlike implicit guidance methods require extensive trial-and-error, we introduce an active action masking mechanism to explicitly prunes unsafe actions from the policy's output distribution.
Our approach dynamically adapts to the environment by using an extended VO method which can handle polyline obstacles by performing intersection checks and distance calculations to identify and mask unsafe headings in real-time. 

For each candidate action representing a desired velocity $\mathbf{v}_i$, we perform parallelized safety checks against both types of hazards: 
For circular obstacles, if $\mathbf{v}_i$ lies within the velocity obstacle cone of any nearby vessel, we calculate the Time-to-Collision (TTC). Any action leading to a TTC below a predefined safety horizon $T_h$ (e.g., $n=5$ control steps) is deemed unsafe and subsequently masked.

For polyline obstacles, we project the vessel's trajectory over the safety horizon $T_h$ to a future position $\mathbf{p}_{\text{next}}$. 
First, we calculate the minimum distance between the vessel's predicted trajectory and the line segment to mask an unsafe action. 
Furthermore, we verify if the vessel's path from its current position $\mathbf{p}$ to $\mathbf{p}_{\text{next}}$ directly intersects with the line segment. If an intersection occurs, the action is also masked. The pseudocode for this mask generation process is detailed in Algorithm 1.

\vskip -0.05in
\begin{algorithm}[H]
    \caption{Active Action Mask Generation (per timestep)}
    \begin{algorithmic}
        \State \textbf{Input:} Current state $s$, candidate headings $\{a_0\dots a_{17}\}$, safety horizon $T_h$
        \State \textbf{Output:} Binary mask $m$
        \State $m[:] \gets 1$ \Comment{Initialize all actions as safe}
        \For{$i$ in $0\dots17$}
            \State $v \gets$ heading\_to\_vel($a_i$, ship\_speed)
            \If{$v \in$ VO\_cone}
                   \If{TTC\_with\_circular\_obstacles($v$) $<$ $T_h$};
                       \State $m[i] \gets 0$; continue
                       \EndIf
            \EndIf
            \If{is\_unsafe\_wrt\_polylines($v$, $T_h$)} 
                \State $m[i] \gets 0$; continue
            \EndIf
        \EndFor
        \State \Return $m$
    \end{algorithmic}
\end{algorithm}
\vskip -0.05in

Concurrently, the final binary mask $\mathbf{m}$ is applied to the logits $z_i$ from the policy network before the softmax operation, producing a safe action distribution for the masked policy $\pi_{\text{masked}}$:

\begin{equation}
\pi_{\text{masked}}(a_i|\mathbf{s}) = \frac{\exp(z_i) \cdot m_i}{\sum_j \exp(z_j) \cdot m_j} .
\end{equation}

This mechanism significantly improves sample efficiency, roughly halving the training steps required for convergence while maintaining low collision rates in congested waters.

\subsection{Domain Randomization}
While the temporal encoding network infers physical dynamics to provide some robustness, a significant gap between simulation and reality persists.
To further bridge this sim-to-real gap, we employ domain randomization. This strategy introduces controlled variability into the simulation to force the policy to learn features that are invariant to minor real-world discrepancies.

In Sim2Sea, we apply minor random perturbations to both sensory observations and command transmissions. More critically, we randomize the ocean current model, as it represents a major source of unmodeled dynamics in the real world. This randomization not only enhances the vessel's adaptability to unpredictable environments but also facilitates our temporal network's ability to recover physical information during training.

Based on the observation that near-shore currents often maintain a stable primary direction over short durations, we model the current with two components: a low-frequency dominant flow and a high-frequency random disturbance. At the start of each training episode, the main direction $\mathbf{d}_{\text{main}}$ and peak amplitude $A$ of the current are randomized. The instantaneous velocity of the current is then expressed as:
\begin{equation}
\mathbf{v}_{current} = A \cdot f(t) \cdot \mathbf{d}_{main} + \epsilon \cdot \mathbf{d}_{random}(t).
\end{equation}
Here, $f(t)$ is a fluctuating multiplier varying in $[0.8, 1.0]$, while the second term represents a minor, time-varying random perturbation $\mathbf{d}_{\text{random}}$ with magnitude $\epsilon$. By exposing the agent to a wide range of such current profiles, we expect it to leverage its temporal network to learn these environmental characteristics and generate appropriate corrective actions, thereby ensuring robust performance upon deployment.

\section{Experiments}

\subsection{Parallel Simulation Performance}

To evaluate the performance of our simulation architecture, we benchmark it by simulating the 3-DOF MMG vessel dynamics using an RK4 integrator. The benchmark involved executing 50 control commands, each comprising 10 physics sub-steps. We measure the average execution time over 10 trials in a large-scale setup consisting of 1024 environments, each containing 64 agents (a total of 65,536 agents).

We compare three parallelization strategies: Full, Per-Environment, and Per-Agent on both consumer and server-grade hardware. Consumer-grade hardware is characterized by higher clock speeds but fewer compute units, while server-grade hardware typically offers more compute units. As shown in Table~\ref{tab:parallel}, our Full Parallelization method is consistently the fastest. The performance gain is most significant on GPUs, where this approach fully utilizes the hardware's parallel processing capabilities. Our design achieves a speedup of over 700-fold on an A100 GPU compared to the slowest baseline (Per-Agent parallelization on CPU), validating its efficiency for large-scale RL training.

\begin{table}[ht]
    \centering
    \caption{Average execution time (in seconds) for simulation with 50 control steps across different parallelization strategies and hardware backends. Lower values indicate better performance. Results are averaged over 10 runs.}
    \resizebox{\linewidth}{!}{ 
    \begin{tabular}{lcccc}
        \toprule
        \multirow{2}{*}{\textbf{Parallelization}} & 
        \multicolumn{2}{c}{\textbf{Windows PC}} & 
        \multicolumn{2}{c}{\textbf{Linux Server}} \\
        \cmidrule(lr){2-3} \cmidrule(lr){4-5} 
        & {{Ryzen 5900X CPU}} & {RTX 5070Ti GPU} & {EPYC 7742 CPU} & {A100 GPU} \\
        \midrule
        Full     & $0.938\pm0.117$  &  $0.036\pm0.002$ & $0.504\pm0.011$ & $0.028\pm0.001$         \\
        Per-Environment & $4.632\pm0.043$ & $0.147\pm0.016$ & $9.667\pm0.037$ & $0.264\pm0.011$  \\
        Per-Agent   & $8.968\pm0.028$ & $1.399\pm0.012$ & $20.368\pm0.194$ & $2.371\pm0.013$ \\
        \bottomrule
    \end{tabular}
    }
    \label{tab:parallel}
    \vskip -0.15in
\end{table}

\subsection{Environmental Setup}

\begin{figure}[ht]
    \centering
    \includegraphics[width=0.9\linewidth]{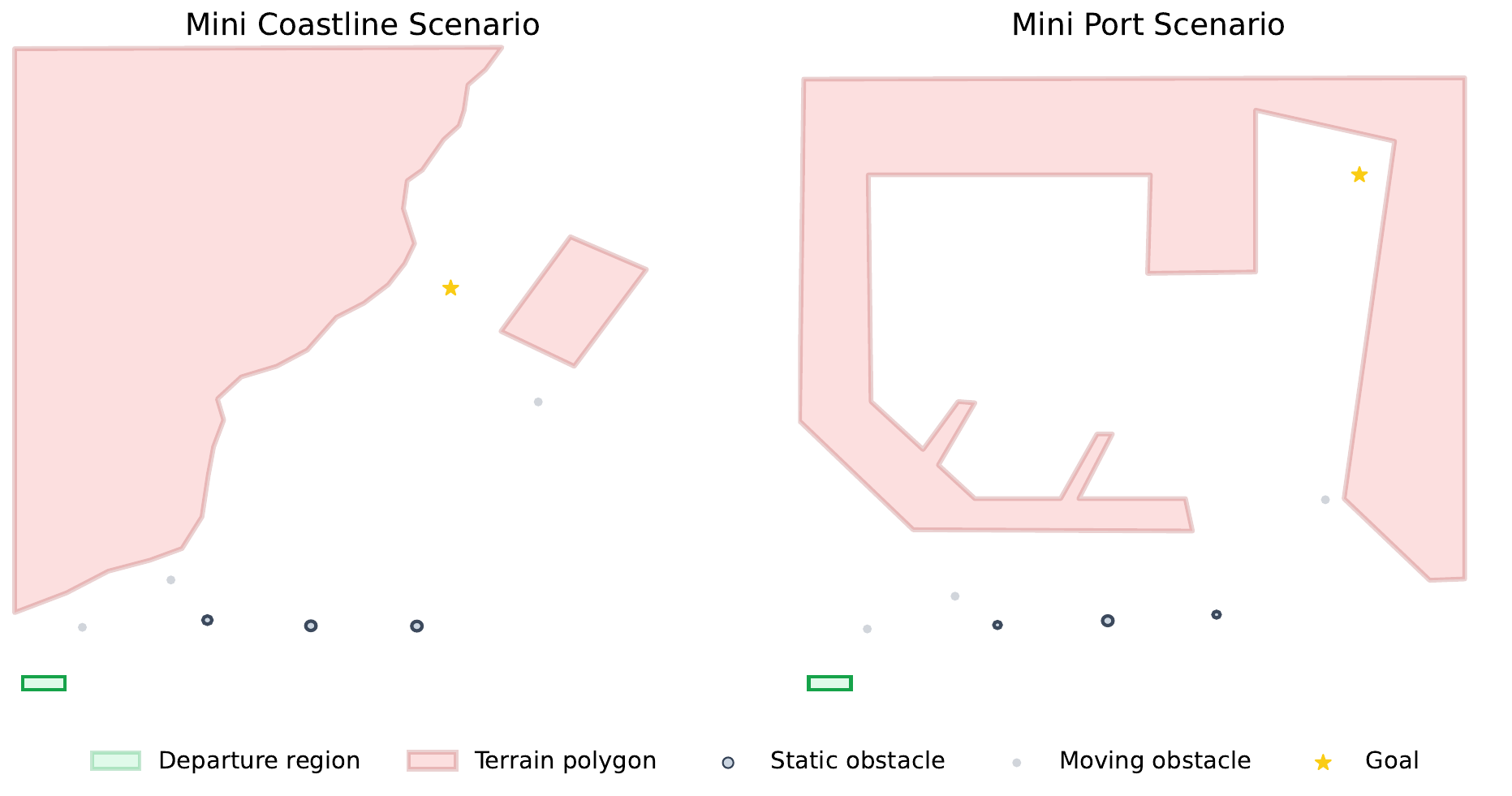} 
    \caption{Mini Coastline and Mini Port scenarios. Pink polygons denote land or restricted terrain, green rectangles indicate the departure regions, yellow stars mark the goals, dark circles are static obstacles, and light circles are moving obstacles.} 
    \label{fig:map}
    \Description{The figure contains two subplots. Left: a coastline with a large landmass and one offshore polygon; a green rectangle at the bottom marks the departure region; a yellow star offshore marks the goal; several dark (static) and light (moving) circular obstacles are scattered along the route. Right: a port enclosed by breakwaters forming narrow fairways and anchorages; a green rectangle near the entrance marks the departure region; a yellow star inside the port marks the goal; multiple static (dark) and moving (light) circular obstacles are placed along the channels. Legend indicates colors and symbols as described above.}
    \vskip -0.15in
\end{figure}

We train and evaluate Sim2Sea in two congested water scenarios: Mini Coastline and Mini Port, depicted in Figure \ref{fig:map}. The height and width of the maps are both 2000 meters. Each episode initializes the agent from a random start area and the task is reaching a fixed goal while avoiding obstacles. We instantiate three static circular obstacles at fixed chart locations and three moving circular obstacles whose initial positions, radii, way-point routes, and speeds are randomized per episode. Moving obstacles travel at constant speed along their way-point loops. The agent dynamics follow the simulation model; obstacle motion is purely kinematic and exogenous for simplification.

\textbf{Mini Coastline}. This near-shore scenario includes a coastline and rectangular aquaculture no-go zones represented as polylines. 

\textbf{Mini Port}. This more challenging scenario requires entering a port through a constrained breakwater opening and reaching an anchorage. 

\textbf{Task protocol}. For both scenarios, an episode terminates when the agent reaches the goal region, CCD detects a collision with obstacles, or the step budget is exceeded. The step budget for mini coastline is 600 steps and is 750 steps for mini port. The reward matches the environment implementation: progress shaping proportional to distance-to-goal improvement, a large positive terminal reward upon success, a large negative penalty upon any collision or boundary infringement, a small per-step time cost, and penalties tied to excessive gradient magnitudes as the agent gets too near to obstacles. 

\subsection{Performance Evaluation in Simulation}

\textbf{Baseline Comparison.} 
Maritime vessel navigation methods commonly rely on COLREGs or Velocity Obstacles. Prior work often adopts reward shaping to encode safety, so we evaluate \textbf{Sim2Sea} against two learning-based baselines that use reward shaping and one classical planner: 
\begin{itemize}
    \item \textbf{VO-RL}: integrates VO in reward shaping.
    \item \textbf{COLREG-RL}: integrates COLREGs in reward shaping.
    \item \textbf{VO}: a pure VO controller.
\end{itemize}
All methods share the same simulator, observation space, discrete action space, and training budget to ensure comparability.

\begin{figure}[ht]
    \centering
    \begin{subfigure}{0.9\linewidth}
        \centering
        \includegraphics[width=\linewidth]{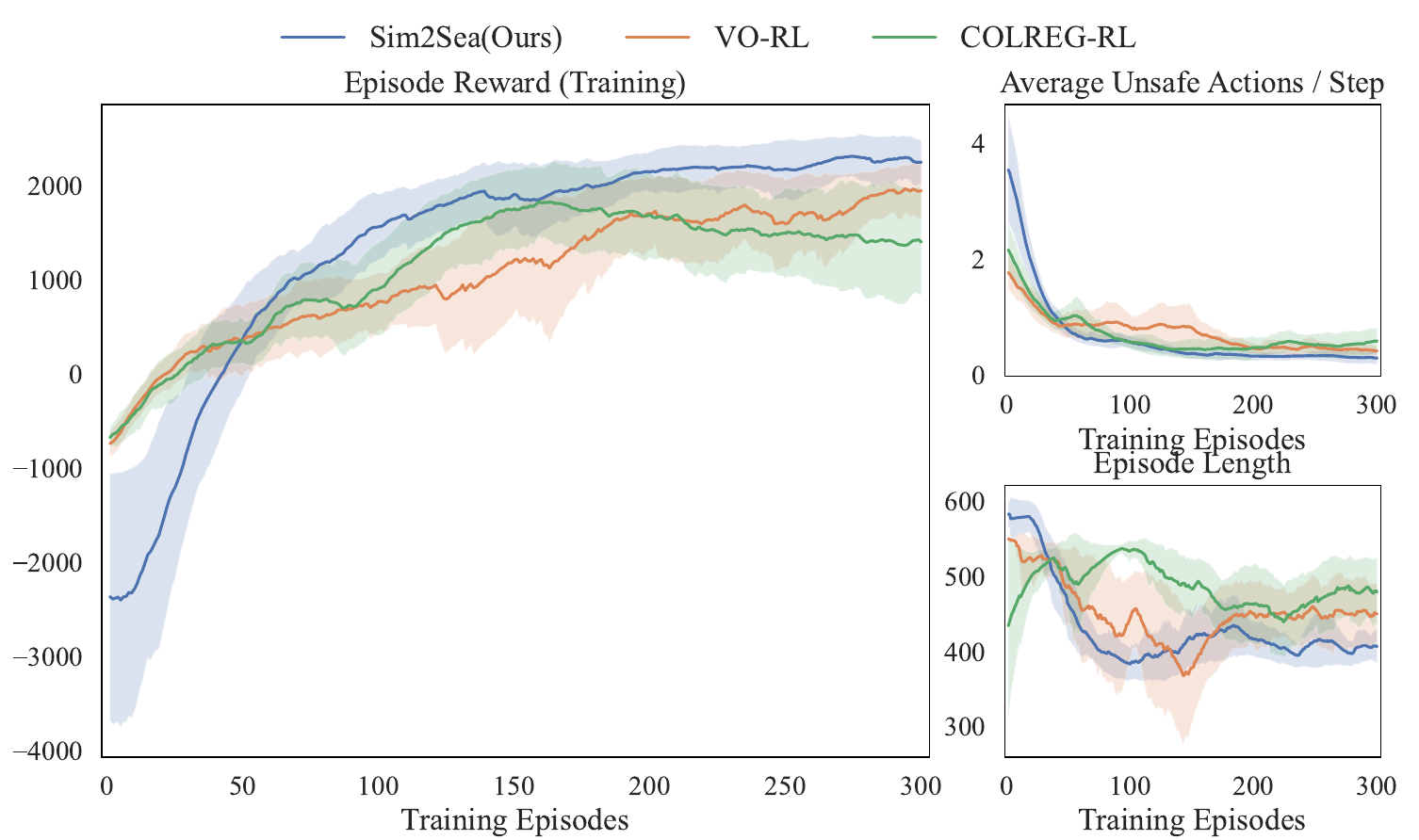}
        \caption{Training curves in the Mini Coastline scenario.}
        \label{fig:minicoast_train}
        \Description{Episode return, average unsafe actions per step, and episode length across training for three variants (ours, w/o mask, w/o BEV) in the Mini Coastline scenario.}
    \end{subfigure}
    \hfill
    \begin{subfigure}{0.9\linewidth}
        \centering
        \includegraphics[width=\linewidth]{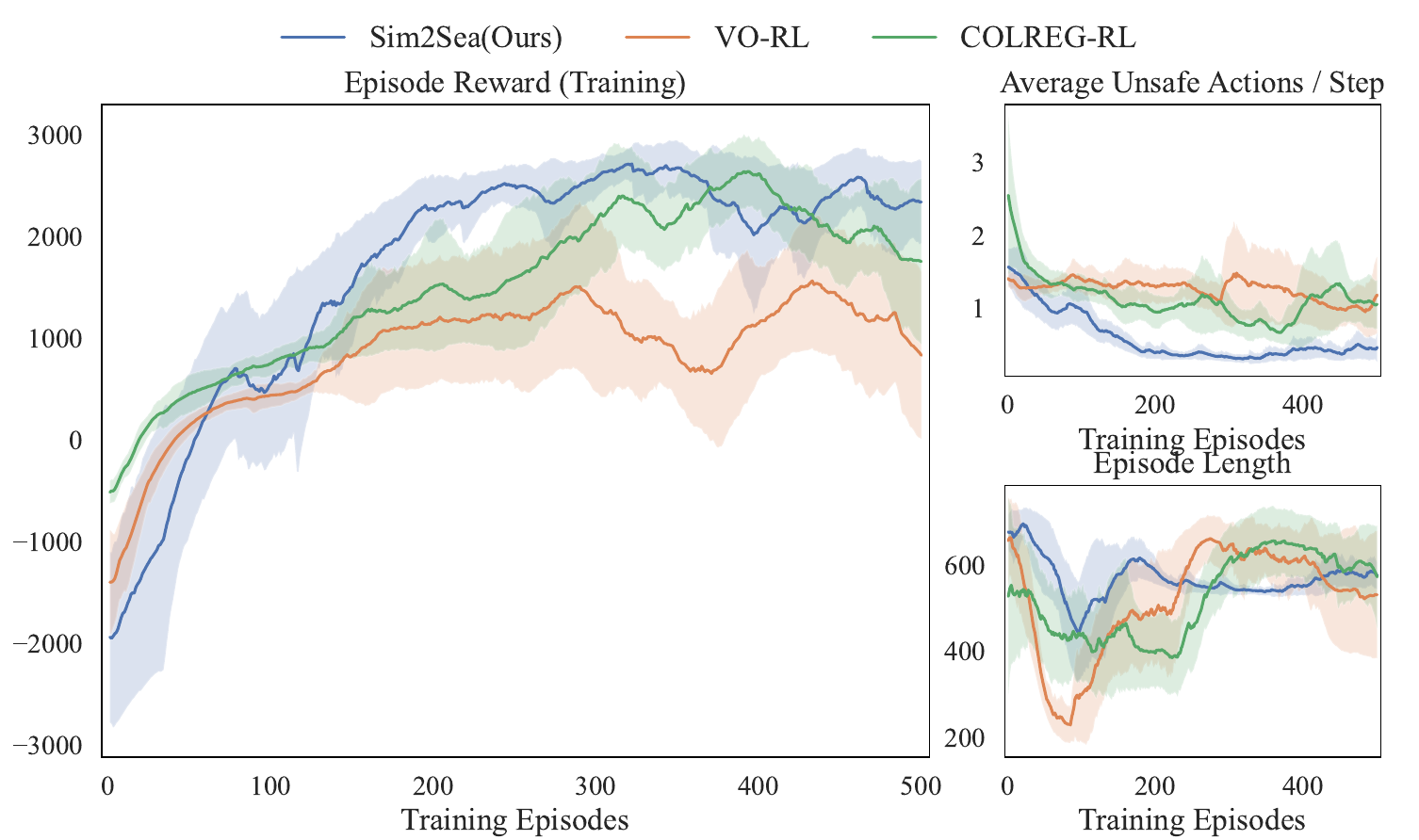}
        \caption{Training curves in the Mini Port scenario.}
        \label{fig:miniport_train}
        \Description{Episode return, average masked actions per step, and episode length across training for three variants (ours, w/o mask, w/o BEV) in the Mini Port scenario.}
    \end{subfigure}
    \caption{Learning performance with and without BEV fusion, active action masking and temporal sequence.}
    \label{fig:baseline}
    \vskip -0.15in
\end{figure}

For VO-RL, we adapt established shaping rules based on Velocity Obstacles~\cite{xie2023drl, li2025dynamic}. When there is no collision risk, the reward penalizes the deviation between the selected action and a VO recommended safe heading, which encourages alignment with safe maneuvers. When risk is present, penalties follow standard VO formulations and increase with intrusion into the VO cone and decreasing time to collision.

For COLREG-RL, we apply similar shaping but emphasize regulatory compliance~\cite{meyer2020colreg}. During collision risk, penalties scale with the distance at closest point of approach (DCPA) and the time to closest point of approach (TCPA), with larger deductions for smaller values that indicate imminent threats. Selecting an action that complies with COLREGs yields a fixed positive reward to reinforce adherence to the rules.


For comparability, we train three seeds per method, select the best checkpoint by validation return, and evaluate ten rollouts per scenario; the VO controller is evaluated ten rollouts. We report success rate, mean episode length, and average unsafe actions encountered per decision step in Table~\ref{tab:inference}.

\begin{table}[ht]
    \vskip -0.05in
    \centering
    \caption{Performance of different methods on MiniCoast and MiniPort Scenarios. SR is success rate in percent. Length is mean episode length in steps. UA is the average number of unsafe actions per decision step.}
    \resizebox{\linewidth}{!}{ 
    \begin{tabular}{lcccccc}
        \toprule
        \multirow{2}{*}{\textbf{Method}} & 
        \multicolumn{3}{c}{\textbf{MiniCoast}} & 
        \multicolumn{3}{c}{\textbf{MiniPort}} \\
        \cmidrule(lr){2-4} \cmidrule(lr){5-7} 
        & {{\textbf{SR(\%)}}} & {\textbf{Length}} & {{\textbf{UA}}} & {\textbf{SR(\%)}} & {{\textbf{Length}}} & {\textbf{UA}} \\
        \midrule
        Sim2Sea     & $93$  &  $500\pm23$ & $1.18\pm0.42$ & $90$  & $343\pm25$ & $0.70\pm0.29$         \\
        VO-RL     & $77$ & $558\pm108$ & $2.24\pm1.43$ & $47$   & $356\pm31$ & $1.28\pm0.37$       \\
        COLREG-RL   & $83$ & $587\pm44$ & $1.96\pm0.32$ & $77$    & $467\pm132$ & $0.92\pm0.34$      \\
        VO   & $70$ & $488\pm22$ & $2.53\pm0.42$ & $67$ & $314\pm9$ & $1.24\pm0.34$       \\
        \bottomrule
    \end{tabular}
    }
    \label{tab:inference}
    \vskip -0.05in
\end{table}

Sim2Sea uses the same primary reward components as VO-RL and COLREG-RL; minor differences in penalty scaling do not alter qualitative conclusions. 
As shown in Figure \ref{fig:baseline} and Table \ref{tab:inference}, Sim2Sea consistently outperforms the baselines. It achieves the highest success rates and lowest average number of unsafe actions by a significant margin. The superior convergence speed and final performance underscore the advantages of employing VO for active action masking over indirect reward shaping. Moreover, while COLREG-RL is competitive, its rule-based rewards are less adaptive than our direct safety mechanism. The pure VO controller yields short paths but its lower success rate reveals the limitations of purely kinematic models that fail to account for vessel dynamics and actuation delays. 

\begin{figure}[ht]
    \centering
    \begin{subfigure}{0.9\linewidth}
        \centering
        \includegraphics[width=\linewidth]{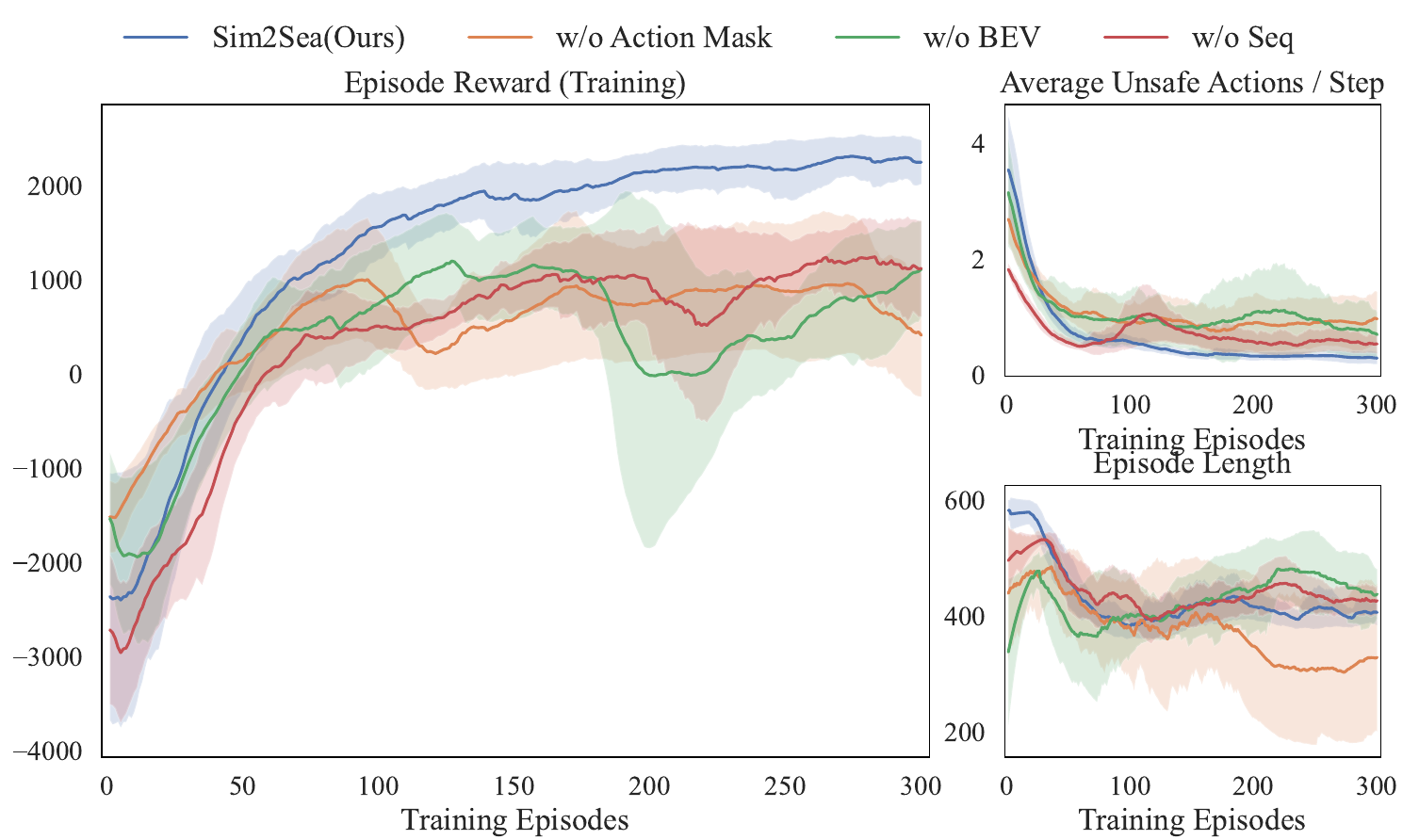}
        \caption{Training curves in the Mini Coastline scenario.}
        \label{fig:minicoast_ablation}
        \Description{Episode return, average masked actions per step, and episode length across training for three variants (ours, w/o mask, w/o BEV) in the Mini Coastline scenario.}
    \end{subfigure}
    \hfill
    \begin{subfigure}{0.9\linewidth}
        \centering
        \includegraphics[width=\linewidth]{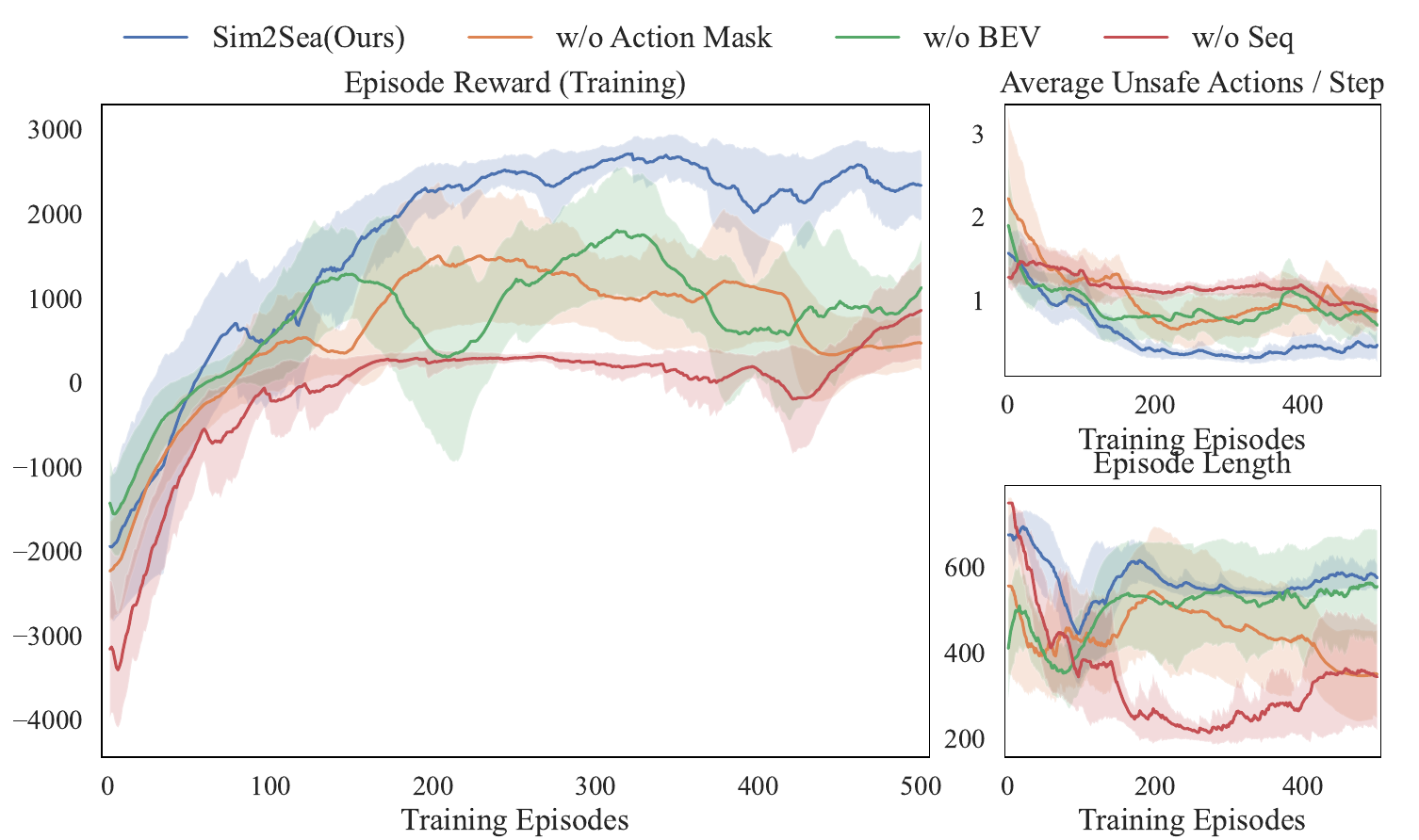}
        \caption{Training curves in the Mini Port scenario.}
        \label{fig:miniport_abalation}
        \Description{Episode return, average masked actions per step, and episode length across training for three variants (ours, w/o mask, w/o BEV) in the Mini Port scenario.}
    \end{subfigure}
    \caption{Learning performance with and without BEV fusion, active action masking and temporal sequence. }
    \label{fig:ablation}
    \vskip -0.15in
\end{figure}

\textbf{Ablation Studies.}
To understand the contribution of each component, we trained three variants of our model: one without action masking, one without the BEV input, and one without the temporal sequence encoder. The results in Figure \ref{fig:ablation} demonstrate that each component is critical for performance.
\begin{itemize}
\item Without action masking (\textbf{w/o Action Mask}): Masking signals are computed but not used to prune actions. The policy samples from the full action set.
\item Without BEV input (\textbf{w/o BEV}): The BEV branch is removed and the CNN encoder is disabled. The policy uses only the temporal encoder over the vectorized features.
\item Without temporal sequence (\textbf{w/o Seq}): The temporal encoder is removed. The policy uses the BEV branch and current step features without history.
\end{itemize}
Each variant is trained on both scenarios with three random seeds and eight parallel environments. We report episode return, average number of masked actions per decision step, and episode length.

As shown in Figure\ref{fig:ablation}, across both scenarios, Sim2Sea consistently achieves higher returns and reaches stable performance in fewer training iterations, which indicates faster convergence. The average unsafe actions per step decreases more rapidly under Sim2Sea, reflecting that the policy remains in safer regions where fewer headings are pruned by VO constraints. This reduction aligns with improved safety because masked actions correspond to headings predicted to violate near term geometric feasibility. Training stability also improves, Sim2Sea exhibits smaller variability across seeds, which suggests that the combination of BEV context and state dependent masking regularizes exploration and produces more predictable learning dynamics.

\subsection{Sim-to-Real Deployment}

\begin{figure}[ht]
    \centering
    \begin{subfigure}{0.9\linewidth}
        \centering
        \includegraphics[width=\linewidth]{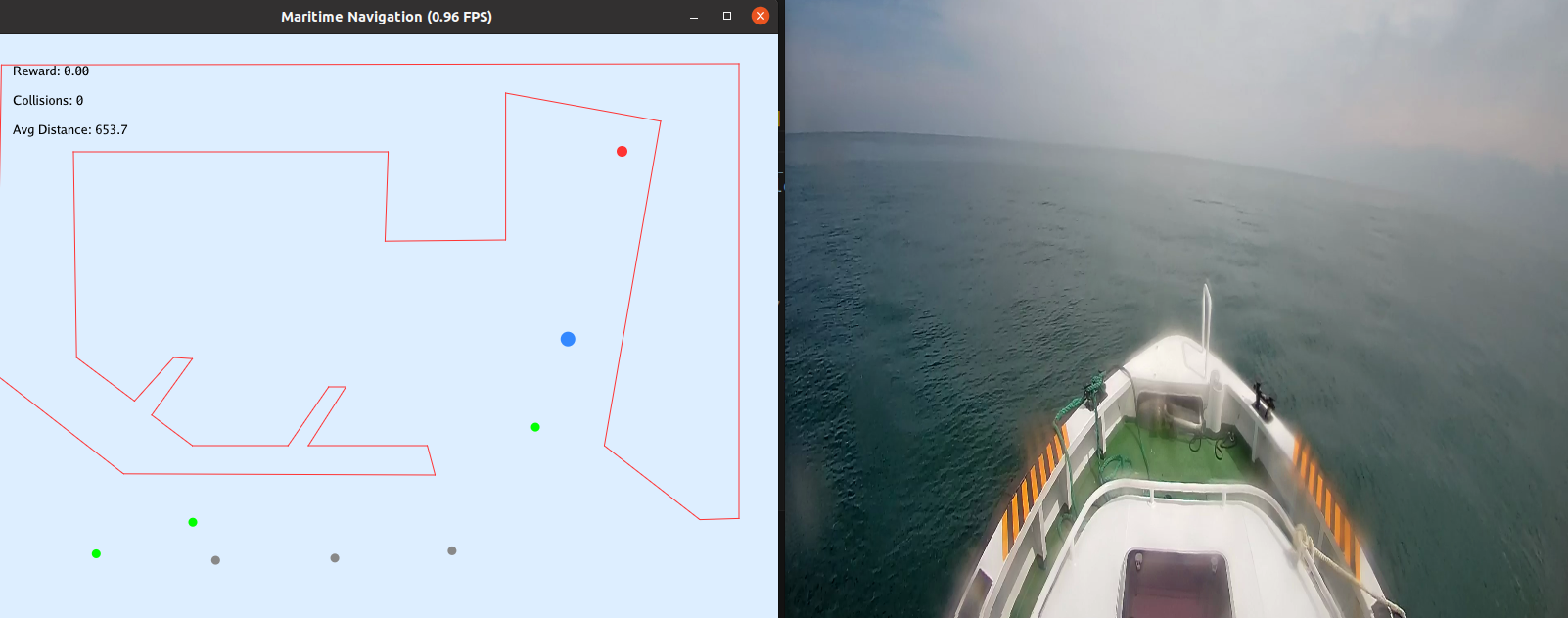}
        \caption{Mini Port scenario with forward camera view.}
        \label{fig:minicoast_ablation}
    \end{subfigure}
    \hfill
    \begin{subfigure}{0.9\linewidth}
        \centering
        \includegraphics[width=\linewidth]{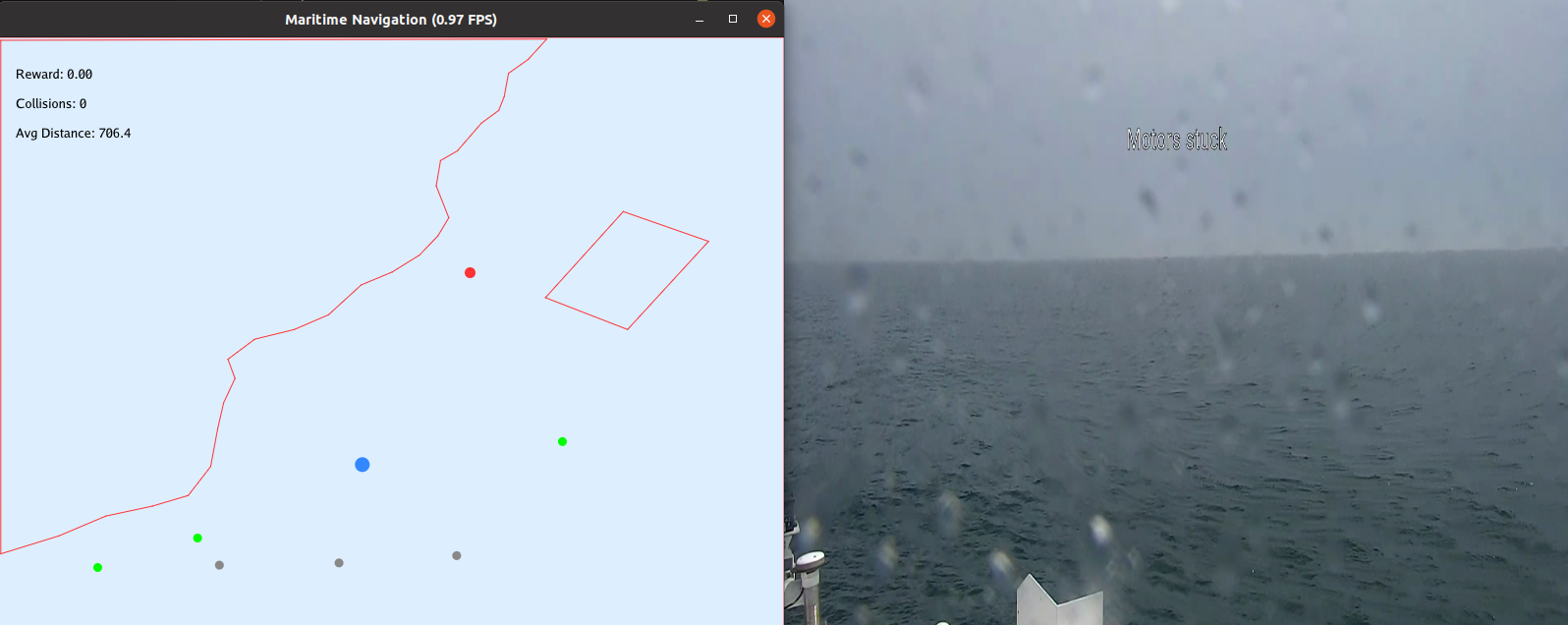}
        \caption{Mini Coastline scenario with side camera view.}
        \label{fig:miniport_abalation}
    \end{subfigure}
    \caption{Onboard interface and camera views. Left: graphical user interface displaying vessel state, mission progress, and chart overlays. Right: forward or side camera view providing situational awareness for safety monitoring.}
    \label{fig:real_scene}
    \vskip -0.15in
\end{figure}

\textbf{Hardware and Onboard System.} 
Sim2Sea is deployed on a 17-ton unmanned surface vessel powered by twin jet propulsion engines with a maximum speed of 32 knots. The platform supports both manual and autonomous operation, and mode switching can be performed at any time to ensure safety. On board computation is provided by a compact Linux server with an Intel Core i7 10700 CPU and an NVIDIA RTX 2080 GPU. Due to power and thermal limits, the autonomy stack is designed for modest resource usage and real-time execution.

The perception suite includes Global Navigation Satellite System (GNSS) for positioning, AIS for cooperative vessel tracking, marine radar for noncooperative targets and obstacles, and forward and lateral cameras for situational awareness. These sensors provide inputs to Sim2Sea and support human supervision. The control loop runs at 1 Hz with a fixed speed of 10 knots and controllable headings. A PID-based low-level controller converts heading and speed commands into throttle and rudder signals. Time-stamped logs are recorded for post hoc analysis.

As shown in Figure~\ref{fig:real_scene}, the onboard computer runs a lightweight graphical interface that visualizes vessel state, mission progress, and chart overlays. Live camera feeds are presented to the safety operator for rapid intervention. A qualified operator remains in the cabin to monitor operations and can switch to manual control whenever risk is detected.


\begin{figure}[ht]
    \centering
    \includegraphics[width=1.0\linewidth]{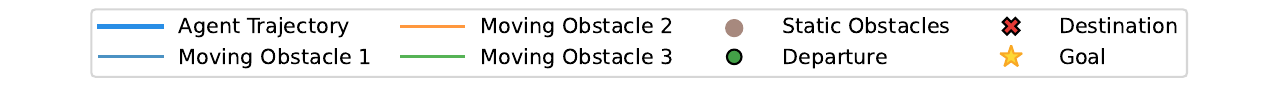}
    \begin{subfigure}{0.3\linewidth}
        \centering
        \includegraphics[width=\linewidth]{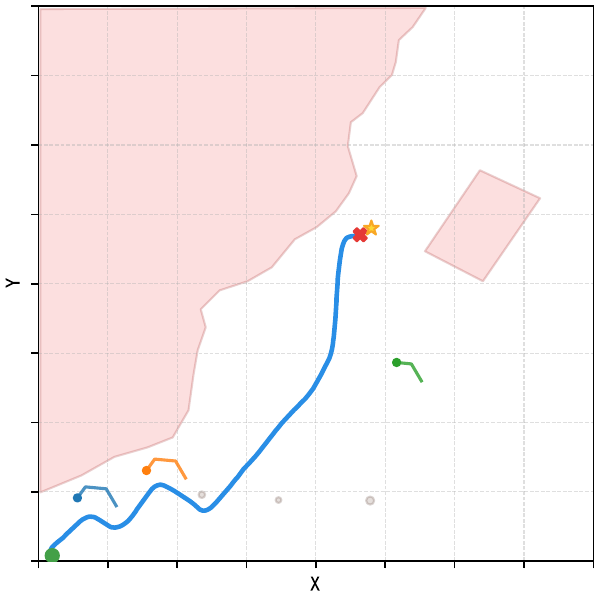}
        \caption{Sim2Sea(Ours):  Trail in Mini Coastline Scenario.}
        \label{fig:traj_ours_coast}
        \Description{}
    \end{subfigure}
    \hfill
    \begin{subfigure}{0.3\linewidth}
        \centering
        \includegraphics[width=\linewidth]{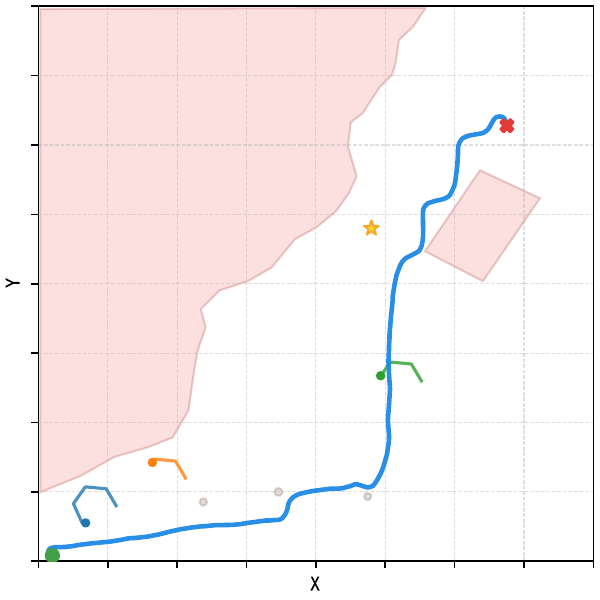}
        \caption{w/o Randomization: Trail in Mini Coastline Scenario.}
        \label{fig:traj_norandom_coast}
        \Description{}
    \end{subfigure}
    \hfill
    \begin{subfigure}{0.3\linewidth}
        \centering
        \includegraphics[width=\linewidth]{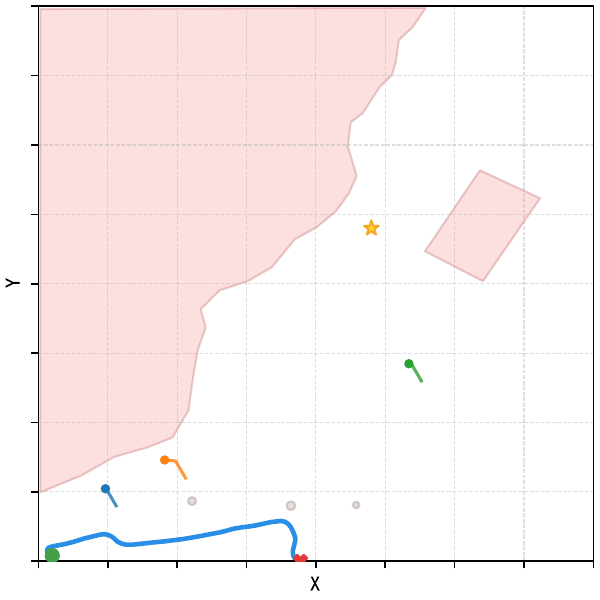}
        \caption{w/o Sequence: Trail in Mini Coastline Scenario.}
        \label{fig:traj_noseq_coast}
        \Description{}
    \end{subfigure}
        \begin{subfigure}{0.3\linewidth}
        \centering
        \includegraphics[width=\linewidth]{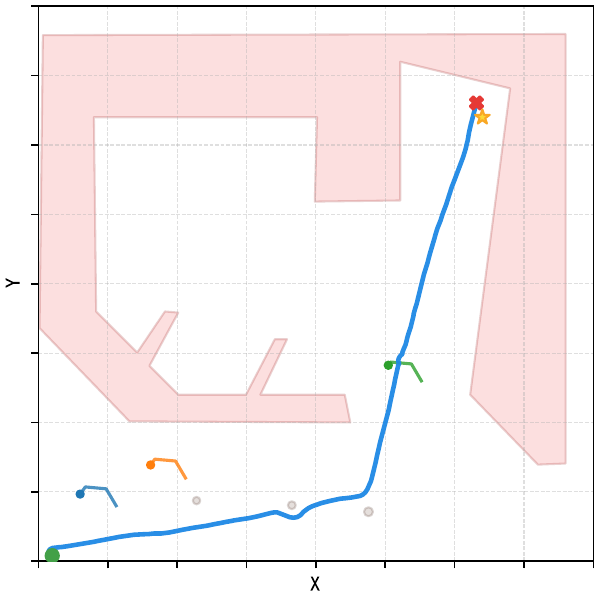}
        \caption{Sim2Sea(Ours): Trail in Mini Port Scenario.}
        \label{fig:traj_ours_port}
        \Description{}
    \end{subfigure}
    \hfill
    \begin{subfigure}{0.3\linewidth}
        \centering
        \includegraphics[width=\linewidth]{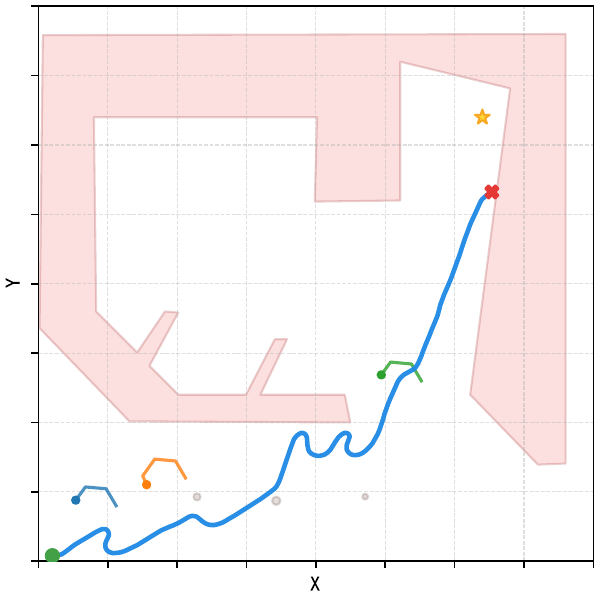}
        \caption{w/o Randomization: Trail in Mini Port Scenario.}
        \label{fig:traj_norandom_port}
        \Description{}
    \end{subfigure}
    \hfill
    \begin{subfigure}{0.3\linewidth}
        \centering
        \includegraphics[width=\linewidth]{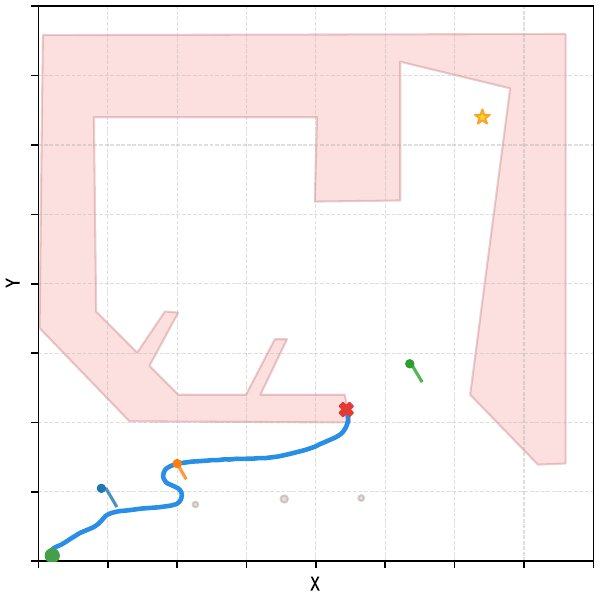}
        \caption{w/o Sequence: Trail in Mini Port Scenario.}
        \label{fig:traj_noseq_port}
        \Description{}
    \end{subfigure}
    \caption{Real-ship sim-to-real trials in two scenarios. Post-convergence trajectories compare Sim2Sea with ablations without domain randomization and without temporal inputs. Obstacles are shown in the figures. The departure is green, the destination is marked with a red cross, and the goal is marked with a star.}
    \label{fig:real_experiments}
    \vskip -0.10in
\end{figure}


\textbf{Zero-Shot Sim-to-Real Transfer Results.}
%
We evaluate the zero-shot deployment of Sim2Sea in open-water trials using the optimal policies trained in simulation. Specifically, we selected the best-performing checkpoints after 300 and 500 training episodes in the Mini Coastline and Mini Port scenarios respectively.
To unify the observation and control frameworks across domains, we align real-world geodetic data with the simulation's metric space by converting GNSS coordinates into a normalized 2D planar system via Mercator projection.
In this planar system, where the positive x-axis represents East and the y-axis North, the vessel's initial position is standardized to the coordinate $[50, 20]$ for all real-world trials.
The deployment environment integrates both virtual and real elements: static obstacles and coastlines follow the simulation map layout, while real-world vessels detected via AIS are dynamically incorporated to evaluate safety.


To isolate the effects of domain randomization and temporal modeling, we compare our \textbf{Sim2Sea} with two variants: a model trained without domain randomization (\textbf{w/o Randomization}), and a model without the temporal encoder that employs an MLP policy (\textbf{w/o Sequence}).



A closer examination of the trajectories in Figure~\ref{fig:real_experiments} reveals the distinct failure modes of the ablated models and underscores the synergistic success of the full Sim2Sea framework. The policy trained without domain randomization (Figures~\ref{fig:traj_norandom_coast},~\ref{fig:traj_norandom_port}) exhibits a relatively brittle behavior; its path is characterized by high-frequency oscillations. This indicates that the policy has overfitted to the idealized dynamics of the simulator and is consequently struggling to handle real-world perturbations
In contrast, the policy without a temporal encoder (Figures \ref{fig:traj_noseq_coast}, \ref{fig:traj_noseq_port}) fails more catastrophically. Its failure is not one of poor navigation but of fundamental incompetence in controlling the vessel. Lacking a temporal model, the reactive MLP policy is incapable of controlling a system with significant inertia, leading to erratic maneuvers and collisions. The VO mask offers only partial protection, as its linear kinematic assumptions are insufficient for complex, non-linear dynamics.

These distinct failures highlight a crucial insight: successful sim-to-real transfer in this domain requires both \textbf{robustness} and \textbf{dynamic awareness}, and these two qualities are not independent but deeply intertwined. The temporal encoder allows the agent to learn an internal model of the vessel's dynamics while domain randomization teaches the agent that this internal model is imperfect and that it must constantly adapt to unmodeled external forces. The synergy of these two components creates a policy that is not just reactive, but predictive and adaptive. Sim2Sea does not simply learn a path; it learns to control a complex dynamic system under uncertainty. 
\section{Conclusion}

In this paper, we introduce Sim2Sea, a sim-to-real maritime vessel navigation framework that integrates a high-throughput simulator, a dual-stream spatiotemporal policy with BEV fusion and Transformer-based temporal encoding, and VO-guided active action masking, to deliver robust decision-making in congested waters. Experimental results demonstrate that Sim2Sea achieves faster convergence, higher success rates, and fewer unsafe actions compared to baseline approaches. 

Notably, Sim2Sea enables successful zero-shot transfer to a real unmanned surface vessel. During deployment, the vessel shows collision-free, smooth, and goal-oriented navigation. Future work will focus on leveraging Sim2Sea’s simulation capabilities and its inherent support for multi-environment and multi-agent settings to pursue more challenging tasks in both simulated training and real-world deployment scenarios.



\vspace{-0.8em}
\begin{acks}
This work was supported by the National Science and Technology Major Project 2022ZD0116404.
\end{acks}

\bibliographystyle{ACM-Reference-Format} 
\bibliography{sample}


\end{document}